\documentclass[10pt,letterpaper]{article}
\usepackage[top=0.85in,left=2.75in,footskip=0.75in,marginparwidth=2in]{geometry}

\usepackage[utf8]{inputenc}

\usepackage{cite}
\usepackage{latexsym}
\usepackage{amssymb}
\usepackage{amsmath}
\usepackage{amsthm}
\usepackage{booktabs}
\usepackage{enumitem}
\usepackage{graphicx}
\usepackage{color}
\usepackage{bm}
\usepackage{acronym} 
\usepackage{verbatim}

\usepackage{nameref,hyperref}

\usepackage[right]{lineno}

\usepackage{microtype}
\DisableLigatures[f]{encoding = *, family = * }

\raggedright
\usepackage{changepage}

\usepackage[aboveskip=1pt,labelfont=bf,labelsep=period,singlelinecheck=off]{caption}

\makeatletter
\renewcommand{\@biblabel}[1]{\quad#1.}
\makeatother

\usepackage{lastpage,fancyhdr,graphicx}
\usepackage{epstopdf}
\fancyheadoffset[L]{2.25in}
\fancyfootoffset[L]{2.25in}

\newcommand{\BibTeX}{B\kern-.05em{\sc i\kern-.025em b}\kern-.08em\TeX}
\DeclareMathOperator*{\argmin}{arg\,min} 

\usepackage{color}

\definecolor{Gray}{gray}{.25}

\usepackage{graphicx}

\usepackage{sidecap}

\usepackage{wrapfig}
\usepackage[pscoord]{eso-pic}
\usepackage[fulladjust]{marginnote}
\reversemarginpar

\begin{document}
\vspace*{0.35in}

\begin{flushleft}
{\Large
\textbf\newline{End-to-end Stroke imaging analysis, using reservoir computing-based effective connectivity, and interpretable Artificial intelligence.}
}
\newline
\\
Wojciech Ciezobka\textsuperscript{1,2},
Joan Falc\'o-Roget\textsuperscript{2},
Cemal Koba\textsuperscript{2},
Alessandro Crimi \textsuperscript{1,*}
\\
\bigskip
\bf{1} AGH University of Krakow,
Krak\'ow, Poland
\\
\bf{2} Sano, center for computational medicine,
Krak\'ow, Poland
\\
\bigskip
* alecrimi@agh.edu.pl

\end{flushleft}

\section*{Abstract}
In this paper, we propose a reservoir computing-based  and directed graph analysis pipeline. The goal of this pipeline is to define an efficient brain representation for connectivity in stroke data derived from magnetic resonance imaging.  Ultimately, this representation is used within a directed graph convolutional architecture and investigated with explainable artificial intelligence (AI) tools.
 
Stroke is one of the leading causes of mortality and morbidity worldwide,  and it demands precise diagnostic tools for timely intervention and improved patient outcomes. Neuroimaging data, with their rich structural and functional information, provide a fertile ground for biomarker discovery. However, the complexity and variability of information flow in the brain requires advanced analysis, especially if we consider the case of disrupted networks as those given by the brain connectome of stroke patients.
To address the needs given by this complex scenario we proposed an end-to-end pipeline. This pipeline begins with reservoir computing causality, to define effective connectivity of the brain. This allows directed graph network representations which have not been fully investigated so far by graph convolutional network classifiers. Indeed,  the pipeline subsequently incorporates a classification module to categorize the effective connectivity (directed graphs) of brain networks of patients versus matched healthy control. The classification led to an area under the curve of 0.69 with the given heterogeneous dataset. Thanks to explainable tools, an interpretation of disrupted networks across the brain networks was possible. This elucidates the effective connectivity biomarker's contribution to stroke classification, fostering insights into disease mechanisms and treatment responses. This transparent analytical framework not only enhances clinical interpretability but also instills confidence in decision-making processes, crucial for translating research findings into clinical practice.

Our proposed machine learning pipeline showcases the potential of reservoir computing to define causality and therefore directed graph networks, which
can in turn be used in a directed graph classifier and explainable analysis of neuroimaging data. This complex analysis aims at improving stroke patient 
 stratification, and can  potentially be used with other brain diseases.


\section*{Introduction}

\label{sec:introduction}
Stroke is one of the leading causes of morbidity and mortality worldwide. Accurate classification can aid in effective treatment and management. Magnetic resonance imaging (MRI) has emerged as a powerful tool for stroke diagnosis, providing detailed images of brain structures and abnormalities. However, the analysis of MRI data poses significant challenges due to its complexity and the need for efficient and reliable classification algorithms, especially when we want to understand the dynamics of the brain.
 
The classification of stroke using medical images has been the primary focus of previous studies \cite{smith2017imaging,stroke-prognostic}. However, most of the approaches carried out so far are focused on the extent of lesions and limited correlation to functional damages such as aphasia and motor deficits \cite{corbetta2018low}. 
Recent studies have started investigating the brain's inner functioning from the point of view of the influence of one brain region on another one, and how lesions compromise those interactions \cite{allegra2021stroke,stroke-prognostic}. Indeed, brain connectivity encompasses the complex interactions between neurons and their intricate network of connections. It is a broad term that encompasses connections between neurons at various levels of granularity and with different connection characteristics. Within this domain, three distinct types of connectivity have emerged: structural (SC), functional (FC), and effective connectivity (EC). Each of these holds clinical and predictive value, offering valuable insights into the brain's intricate workings \cite{sporns2013structure}. Effective connectivity investigates the causal link between the time series of two regions of the brain and can be represented as directed graphs. Classification and explanation of directed graphs have not been fully investigated and the study of stroke with those tools provides the opportunity to create a pipeline exploring all those elements.

More specifically,  local ischemia damages neurons and structural neural connections at the site of injury. This affects primarily subcortical regions, subsequently altering long-range functional connectivity between cortical areas. Decreases in functional connectivity alterations suggest deficits but cannot reveal the directionality or time scale of the information flow, leaving several open questions related to the directionality and functioning of the brain after a non-traumatic injury such as a stroke. 
Allegra and colleagues carried out previous studies where this transfer of 
information view of the brain of stroke patients was investigated through 
Granger Causality (GC) analyses \cite{allegra2021stroke}, where they observed a significant decrease in inter-hemispheric information transfer in stroke patients compared to matched healthy controls. GC has been used largely in computational neuroscience studies due to its low computational costs compared to other methods \cite{friston2003dynamic,ccm-original}. Practically, the method estimates autoregressor variables relating to different time series which are then further validated by F-statistics to establish causality. Yet, due to the potential confounding characteristics that each autoregressor may generate  \cite{maziarz2015review}), there are still ongoing disagreements on whether this can help define causal interaction between brain regions \cite{reid2019advancing} using this framework, and some authors consider GC as just a relation measures \cite{etkin2018addressing}. 
To overcome these limitations, researchers have explored the use of reservoir computing in a completely detached paradigm to extract causality \cite{rcc-original,wholebrainRCC}. 
Reservoir computing is a computational framework that leverages the dynamics of recurrent neural networks to process and classify complex temporal data effectively, by exploiting the inherent memory and non-linear dynamics of reservoirs \cite{esn-original,lsm-original}. It has also been used to classify electroencephalography data from stroke patients \cite{bouazizi2024novel}, though as a classifier itself, not to estimate the structure of the human brain. 

Finally, capturing both spatial and temporal patterns can help understand stroke beyond traditional voxel-based lesion-symptom mapping \cite{bates2003voxel} to consider specific information transfer and interactions in the brain \cite{fox2018mapping,wholebrainRCC}. Technically, this will produce a directed graph representation that can be classified and explored with explainable AI tools.  
 
\begin{figure*}[h!]
\centering
\includegraphics[width=0.9\linewidth]{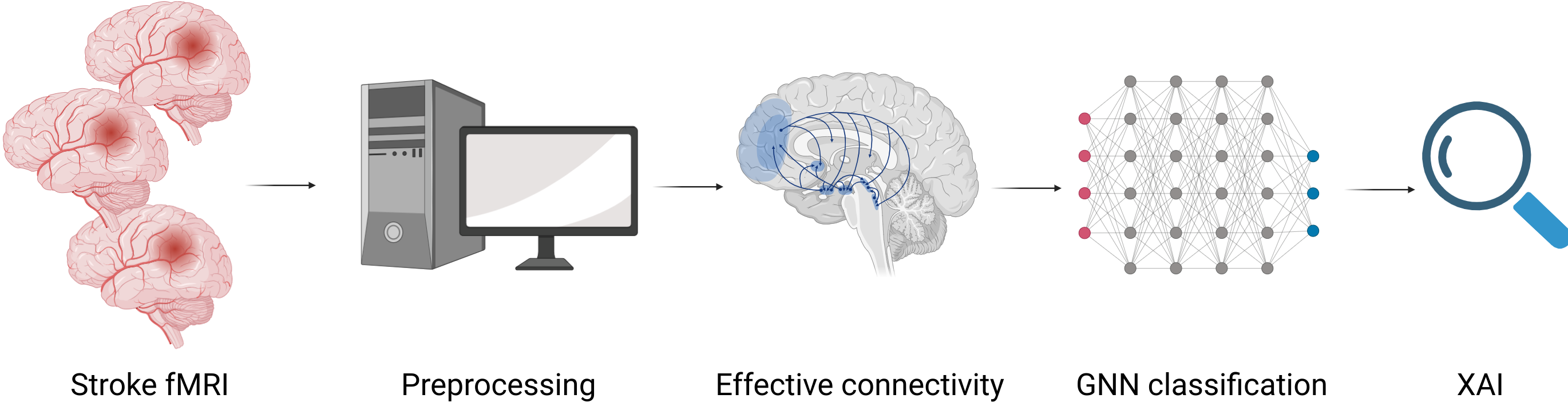}
\caption{Overall pipeline of the study, where MRI data are preprocessed, used to define an effective connectivity representation, classified and the results are investigated by 
 explainable AI tools.}
\label{fig:pipeline}
\end{figure*}

In summary, using reservoir computing we i) defined causality in stroke patients, and, given the generated representation of causality as directed graphs, investigated ii) the value of the resulting directed maps together with their classification, and iii) the explainability of the classification to provide insights into the overall brain network disruption in stroke patients (Fig. \ref{fig:pipeline}). To our knowledge, no study has classified directed graphs and explained their significance in computational neuroscience and neurology. Thus, incorporating these features into classification algorithms could improve stroke diagnosis accuracy and efficiency.

\section{Methods}
\subsection{Data and preprocessing}
The dataset was previously collected by the School of Medicine of the Washington University in St. Louis and complete procedures can be found in \cite{corbetta2015common}. They collected MRI data and behavioral examinations of stroke patients and healthy controls. The imaging data comprise structural and functional MRI from controls and patients suffering from hemorrhagic and ischemic stroke. Acquisitions were done within the first two weeks of the stroke onset (i.e., acute). Structural scans include T1-weighted, T2-weighted, and diffusion tensor images. Functional images include a resting state paradigm. Scanning was performed with a Siemens 3T Tim-Trio scanner. Briefly, we closely followed the pre-processing steps outlined in \cite{Koba2024}. Following a quality control of \emph{fMRIPrep} outputs, 104 stroke subjects and 26 control subjects were qualified for further analysis. 
For our purposes, it suffices to say that structural scans were used in combination with functional acquisitions to co-register all participants into a common template. Gray matter signal was finally obtained after artifact removal and parcellated into 100 regions of interest (ROIS) \cite{fmriprep,schaefer-parcellation}. For every subject and patient, these 100 time series (i.e., one for each ROI) were fed into the pipeline outlined below to obtain the subject-specific effective connectivity maps.

The dataset is not public but it is available upon request
 to the original authors \cite{corbetta2015common}. 
The used code is instead available at the URL 
\url{https://github.com/Wotaker/Effective-Connectivity-Reservoir-Computing}.

\subsection{Reservoir computing}
\label{sec:rc}

Reservoir computing networks (RCN), despite being known for more than two decades, have been largely eclipsed by other frameworks. A reservoir network is a set of artificial neurons that are randomly connected between themselves thus forming a recurrent architecture \cite{esn-original,lsm-original}. 
Sometimes this is also called \textit{echo-state network} since the internal dynamics of the reservoir (or "echo state") maintain information about the system's input history.  In this framework, an input series $\mathbf{u}_t$ is fed into this high dimensional dynamical system of $N$ units through a non-linear activation function,

\begin{equation} \label{eq:RC_input}
    \mathbf{r}^{in}_t = f^{in}(\mathbf{W}^{in} \mathbf{u}_t),
\end{equation} where $\mathbf{W}^{in}$ is an $N$ x $(N_{in}+1)$ matrix of random weights including biases, $N_{in}$ is the dimensionality of the multivariate input, and $f_{in}$ is the non-linearity. At each time step $t$ the former projection is used to drive the reservoir units $\mathbf{r}_t$. The current state of each unit is a combination of the past states as well as the current input,
\begin{equation} \label{eq:RC-rec}
    \mathbf{r}_t = (1-\lambda) \mathbf{r}_{t-1} + \lambda f (\mathbf{r}^{in}_t + \mathbf{W} \mathbf{r}_{t-1}),
\end{equation} where $\mathbf{W}$ is an $N$ x $N$ matrix of random weights, and $\lambda$ is the leakage that controls the importance of the reservoir's history to the current time stamp $t$. The final component of the reservoir is a set of \textit{readout} weights $\mathbf{W}^{out}$ that extract information from the hidden state and map onto specific predictions. That is,

\begin{equation}
\label{eq:RC_output}
\mathbf{y}_t = \mathbf{W}^{out} \mathbf{r}_t.
\end{equation} The predictions $\mathbf{y}_t$ might be of arbitrary dimension $N^{out}$ and, importantly, are linear w.r.t. to the reservoir states. Within this paradigm, only that readout weights $\mathbf{W}^{out}$ are trained via incremental linear regression optimization \cite{pyrcn,liang2006fast},

\begin{equation} \label{eq:lin-reg}
\mathbf{W}^{out} = (\mathbf{R} \mathbf{R}^T + \alpha \mathbf{I})^{-1} (\mathbf{Y} \mathbf{R}^T),
\end{equation} with $\alpha$ being a regularization parameter, $\mathbf{R}$ is the matrix obtained after concatenating all the reservoir states, and $\mathbf{Y}$ contains the outputs. Once again, the readout weights contain a set of $N_{out}$ biases.

Noteworthy, as opposed to other architectures suited for time series forecasting, only a reduced set of output weights needs to be trained, thus increasing its computational efficiency. The random weights $\mathbf{W}^{in}$ are drawn from a uniform distribution bounded between -1 and 1. The recurrent connections are drawn from a standard normal distribution and are later scaled by the spectral radius. The latter largely ensures that the network possesses the echo-state property, although there is recent evidence 
disagreeing with this aspect \cite{yildiz2012re,hands-on-rc}. 

Briefly, the main idea behind reservoir-like computing is that a given input pushes the reservoir to specific locations in a high-dimensional manifold \cite{esn-original,lsm-original}; the output weights are then optimized to retrieve information from the nearby regions. Were the input to move the reservoir away to other points, the output weights would not be able to recover meaningful information hence completely missing the prediction. Further evidence suggests that RCNs supersede deep learning-based models for temporal series prediction even on the verge of chaos \cite{shahi2022prediction}. Richer approaches aim to train the reservoir connections themselves and have been proven to be useful in understanding the dynamical properties of cortical networks \cite{Parga_2023}, offering an interesting framework for similar use cases.
The parameter values used in our experiments can be found in Table \ref{tab:RCN-parameters}. 

\begin{table}
    \centering
    \begin{tabular}{|c|cc|}
    \hline
    \textbf{Object}            & Input-to-Node & Node-to-Node   \\ \hline
    \textbf{Units} (\#)        & 50            & 50             \\
    \textbf{Sparsity}          & 1             & 1              \\
    \textbf{Activation}        & logistic      & tanh           \\
    \textbf{Scaling}           & 1             & \textit{NA}    \\
    \textbf{Shift}             & 0             & \textit{NA}    \\
    \textbf{Bias scaling}      & 1             & \textit{NA}    \\
    \textbf{Bias shift}        & 0             & \textit{NA}    \\
    \textbf{Random seed}       & \textit{null} & \textit{null}  \\
    \textbf{Spectral radius}   & \textit{NA}   & 1              \\
    \textbf{Leakage} ($\lambda$) & \textit{NA} & 1              \\
    \textbf{Bidirectional}     & \textit{NA}   & \textit{false} \\
    \hline
    \end{tabular}
    \caption{Summary of the parameters chosen to train the Reservoir Computing Networks (RCNs) in this work. For the two different blocks, \textit{NA} stands for Not Applicable, and \textit{null} indicates that the value was left empty to be chosen by the implemented random sampler. For further details on the meaning of each one of these parameters we refer the reader to the original publication of the package \cite{pyrcn} and documentation.}
    \label{tab:RCN-parameters}
\end{table}

\subsection{Reservoir computing networks to map causal interactions in lesioned brains}
\label{sec:rcc}

Traditionally, effective connectivity in neuroimaging can be estimated in different ways, as dynamic causal modeling \cite{friston2003dynamic}, GC \cite{granger-original}, continuous-time implementations \cite{gilson2016estimation}, or information theory \cite{vicente2011transfer}. 
Granger-like interpretations are often preferred due to their relative computational costs and implementation, though they are
not exempt from controversy \cite{grassmann2020new} thus justifying alternative approaches. 

An unrelated proposal relies on the properties of the state-space of the dynamical system to reconstruct asymmetric mappings between delayed embeddings of each component of the system \cite{ccm-original}. That is, it leverages Taken's theorem to find the optimal neighborhood as well as the exact delay at which the reconstruction is optimal. Recent extensions \cite{ccm-phase-space-prediction,eccm-original,breston2021convergent} have incorporated non-linear methods as well as reducing the number of ad-hoc parameters. Most prominently, reservoir computing has proven to be an efficient and accurate alternative to automatize the process almost in its entirety \cite{rcc-original}. 

Let's consider the relationship between two one-dimensional variables, $x$ and $y$, where it hypothesizes that the delay at which interactions take place is not smaller than the sampling rate (e.g., Time of Repetition in functional MRI). The prediction skill, denoted by $\rho_{x \rightarrow y}(\tau)$, is defined as the Pearson correlation between the true time series, $\mathbf{y}(t+\tau)$, and the predicted series $\hat{\mathbf{y}}(t+\tau)$ from the input $\mathbf{x}(t)$.
\begin{equation} \label{eq:prediction-skill}
\rho_{x \rightarrow y}(\tau) := \text{corr }\left[\mathbf{y}(t+\tau), \hat{\mathbf{y}}(t+\tau)\right].
\end{equation} 

Noteworthy, the Pearson correlation between the true and reconstructed series ($\rho$) is used  to estimate directedness, though other metrics like mean squared error could also be used. Directionality can still be assessed using the same hypothesis testing mechanisms \cite{ccm-phase-space-prediction}.

Moreover, the time series are fed into the reservoir \textit{all-at-once}, letting the network project all of them. The neighboring points in the variable's embedding are then remapped to the target embedding via the training of the output weights. It should noted that this represents a deviation from more canonical usages \cite{rcc-original,hands-on-rc}. To investigate the causal relationship between variables, we first calculate both $\rho_{x \rightarrow y}(\tau)$ and $\rho_{y \rightarrow x}(\tau)$ in a given temporal domain. We then examine the values of $\tau$ at which either $\rho_{x \rightarrow y}(\tau)$ or $\rho_{y \rightarrow x}(\tau)$ reaches its peak value \cite{eccm-original,rcc-original}. To streamline the subsequent description, we introduce the following notation:
\begin{equation} \label{eq:tau-notation}
\begin{aligned}
\tau_{x \rightarrow y} &:= \argmin_{\tau} \rho_{x \rightarrow y}(\tau) \\
\tau_{y \rightarrow x} &:= \argmin_{\tau} \rho_{y \rightarrow x}(\tau).
\end{aligned}
\end{equation} Empirically, directionality is then defined as follows \cite{ccm-original}:

\begin{itemize}
    \item if $\tau_{x \rightarrow y}$ is positive, and $\tau_{y \rightarrow x}$ is negative, we say that $x$ causes $y$;
    \item if $\tau_{x \rightarrow y}$ is negative, and $\tau_{y \rightarrow x}$ is positive, we say that $y$ causes $x$;
    \item if both $\tau_{x \rightarrow y}$ and $\tau_{y \rightarrow x}$ are negative, we say that $x$ and $y$ causes each other.
\end{itemize}

Despite seeming counterintuitive, information of $\mathbf{y}$ is present in earlier observations of $\mathbf{x}$ and, consequently, that current information of the cause $\mathbf{x}$ is useful to predict future observations of the consequence $\mathbf{y}$ (see \cite{ccm-original} for a comprehensive explanation). In certain systems, predictability scores peak at negative lags $\tau<0$ for both directions, being the height of the peaks informative of the coupling strength \cite{rcc-original}. However, the existence of this bidirectionality does not necessarily invalidate the former statements \cite{wholebrainRCC}.
 
\begin{figure}[h]
\centering
\includegraphics[width=\linewidth]{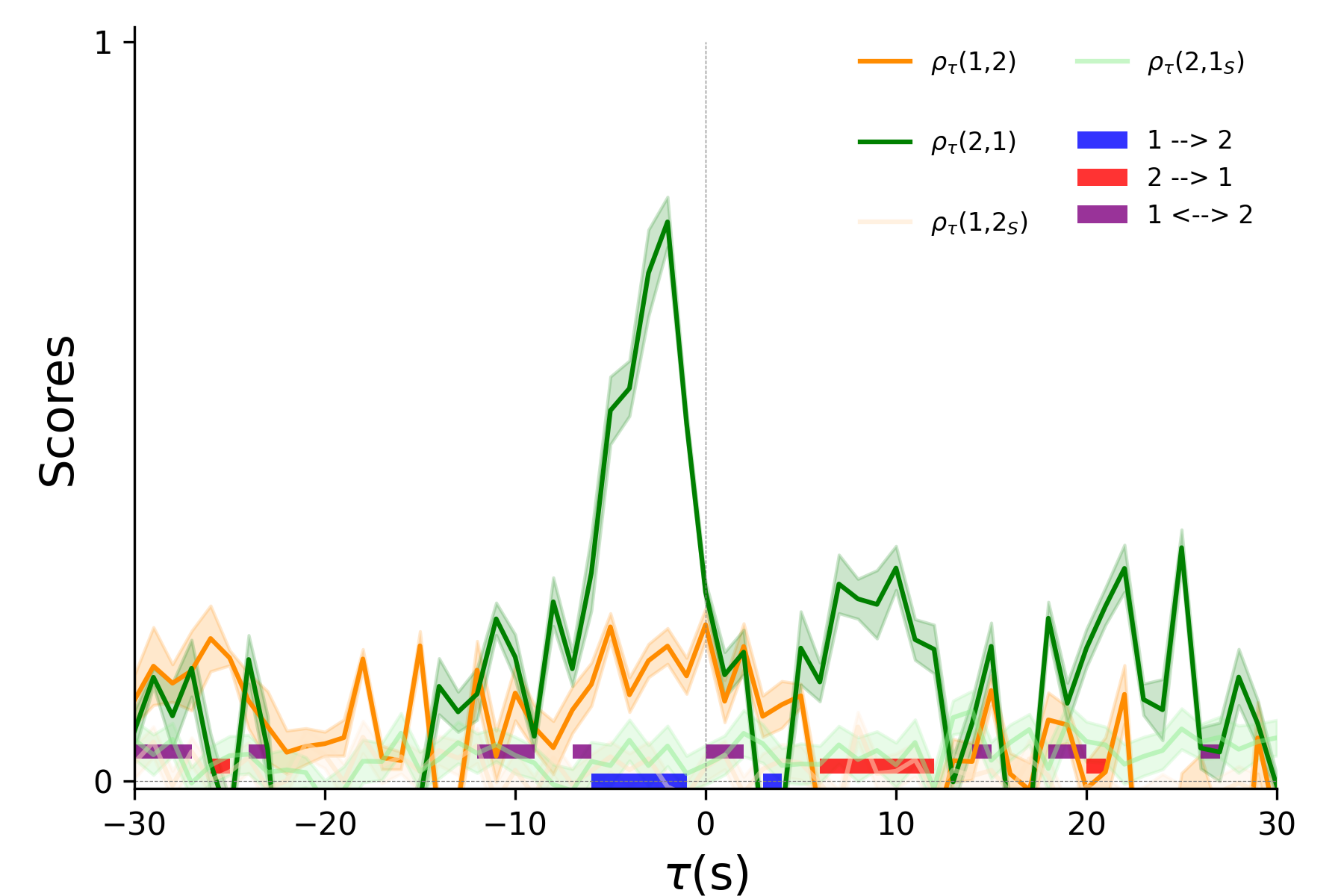}
\caption{Predictability scores from an the same chaotic system defined in \cite{ccm-original,wholebrainRCC}. Solid lines show the predictability in Eq. \ref{eq:prediction-skill} between embeddings. Shaded regions show 1 standard error of the mean. Transparent lines show the predictability of the surrogate system, which is used to define the expected level of chance against which the hypotheses are tested. In this academic example, it can be said that strong asymmetric interactions between two time series exist at different temporal lags.}
\label{fig:example}
\end{figure}

It was quickly noted that in large and noisy networks, such as the brain, it is unlikely that the predictability scores in Eq. \ref{eq:prediction-skill} reach clear and distinct peaks. Functional signals are notoriously noisy \cite{fmri-networks}, and indeed prediction with this approach is challenging  \cite{avvaru2023effective}. A solution to this issue relies on assessing the minimal requirements that are needed to suggest causal interactions \cite{wholebrainRCC}. For that, the difference between prediction scores should be evaluated and contrasted with proper surrogate predictions \cite{mccracken2014,iaaft-1,iaaft-2}. That is,

\begin{equation}
\label{eq:big-delta}
\Delta_{x \rightarrow y}(\tau) := \rho_{x \rightarrow y}(\tau) - \rho_{y \rightarrow x}(\tau),  
\end{equation} which can be interpreted as an indication of the potential causality direction (Table \ref{tab:delta-score}). The scores in Eqs. \ref{eq:prediction-skill} and \ref{eq:big-delta} can be contrasted against the 95\% confidence interval obtained from a surrogate testing procedure \cite{wholebrainRCC}. It has been shown that the requirements to define causality can be compressed into a reduced set of $\delta$-scores \cite{wholebrainRCC},

\begin{table}[t]
    \centering
    \begin{tabular}{c|cc}
         & $\tau > 0$ & $\tau < 0$\\
         \hline
         $\Delta_{x \rightarrow y}(\tau) > 0$ & $x \rightarrow y$ & $y \rightarrow  x$\\
         $\Delta_{x \rightarrow y}(\tau) < 0$ & $y \rightarrow x$ & $x \rightarrow  y$\\
    \end{tabular}
    \caption{Potential causal directions based on the sign of $\Delta$-score and the positive or negative $\tau$ regime.}
    \label{tab:delta-score}
\end{table}

\begin{equation} \label{eq:small-delta-uni}
\delta_{x \rightarrow y}(\tau) := \begin{cases}
        (1-p_{\rho_{x \rightarrow y}(\tau) > 0})(1-p_{\Delta_{x \rightarrow y}(\tau) > 0}) & \text{if $\tau>0$}\\
        (1-p_{\rho_{y \rightarrow x}(\tau) > 0})(1-p_{\Delta_{y \rightarrow x}(\tau) > 0}) & \text{if $\tau<0$}
    \end{cases}
\end{equation} for directed interactions, and

\begin{equation} \label{eq:small-delta-bi}
\delta_{y \leftrightarrow x}(\tau) := (1-p_{\rho_{x \rightarrow y}(\tau) > 0})(1-p_{\rho_{y \rightarrow x}(\tau) > 0})p_{\Delta_{x \rightarrow y} \neq 0}
\end{equation} for bidirectional interactions. $p_{H_1}$ is the $p$\nobreakdash-value after testing the alternative hypothesis $H_1$ against the surrogate data (Fig. \ref{fig:example}). For instance, $p_{\rho_{x \to y}}$ is a p-value for the hypothesis that x influences y. The values of the $\delta$\nobreakdash-scores range from 0 to 1, with higher values indicating greater confidence in the existence of a causal relationship with a coupling delay of $\tau$ between the examined variables.

Then, for a given lag $\tau$, a matrix $\mathbf{A}_\tau$ collects the $\delta$\nobreakdash-scores, where each element $[x,y]$ represents the causal relationship from node signal x to ROIs signal y,
\begin{equation}
\label{eq:ec-matrix}
\bm{A}_\tau[x, y] = \begin{cases}
    \delta_{x \rightarrow y}(\tau) & \text{unidirectional}\\
    \delta_{x \rightarrow y}(\tau) + \delta_{x \leftrightarrow y}(\tau) & \text{bidirectional.}\\
\end{cases}
\end{equation} The effective connectivity (RC) matrix $\mathbf{A}_\tau$ is a final representation of the effective connectivity network of every subject; it is directed, non-symmetric, and can incorporate bidirectional causality connections. For our experiments, for every possible interaction $x \to y$, we trained 20 different reservoirs and tested against 100 shuffled targets, strictly following what was outlined in \cite{wholebrainRCC}. Furthermore, only unidirectional connections were kept from the adjacency matrix in Eq. \ref{eq:ec-matrix}. 

In our experiments, we investigated the classification of pathological groups with the effective connectivity matrices used as features (Fig. \ref{fig:architecture} TOP), and we also compared those to the effective connectivity matrices obtained by Granger causality, representing one of the state-of-art approaches. As a last step, for each entry $A_{\tau}\left[x,y\right]$, we standardized all samples by subtracting the mean connectivity of the control group and dividing by the standard deviation. Finally, these standardized causal relationships (i.e., directed graphs) were fed into two simple graph classifiers to explore and explain the most informative nodes and links to detect stroke occurrence. 

\subsection{Graph convolutional neural networks}
Graph convolutional neural networks (GNNs) are a variation of traditional convolutional neural networks which capitalize on graph data representations
and can learn non-trivial representations by leveraging the complex topological organization of the data \cite{velivckovic2023everything}. Intuitively, a graph constitutes a non-Euclidean geometric space where complex relationships between data points can be embedded and forwarded as inputs into a GNN \cite{BronsteinGeometricDL}. More formally, a graph $\mathcal{G} = (\mathcal{V}, \mathcal{E})$ is defined as a set of nodes $\mathcal{V} = \{1, \ldots, n\}$ and a set of edges $\mathcal{E} = \{(i,j) \ | \  i,j \in \mathcal{V}\}$ where $(i,j)$ represents a link or interaction between the $i$-th and $j$-th nodes. Initially, each node $i \in \mathcal{V}$ is associated with a column feature vector $\mathbf{h}_{i}^{(0)} \in \mathbb{R}^{d^{(0)}}$.

Every layer $l$ of a GNN updates the hidden representation of each node by aggregating information from the neighborhoods: 
\begin{equation} \label{layer}
    \mathbf{h}_{i}^{(l+1)} = f_{\theta} \left( \mathbf{h}_i^{(l)}, \textsc{F} \left( \{ \mathbf{h}^{(l)}_j \, | \, j \in \mathcal{N}_i \} \right) \right),
\end{equation}
where $\textbf{h}_{i}^{(l+1)} \in \mathbb{R}^{d^{(l+1)}}$ are the new node representations, $\mathcal{N}_i$ is the neighborhood of the $i$-th node, $f_{\theta}$ denotes a nonlinearity, and $\textsc{F}$ is a permutation-invariant aggregator. Several proposals exist for the aggregation operator, determining the expressive power, interpretability, learning stability, and scalability of the network \cite{velivckovic2023everything}. 

\begin{figure*}[h]
\centering
\includegraphics[width=\linewidth]{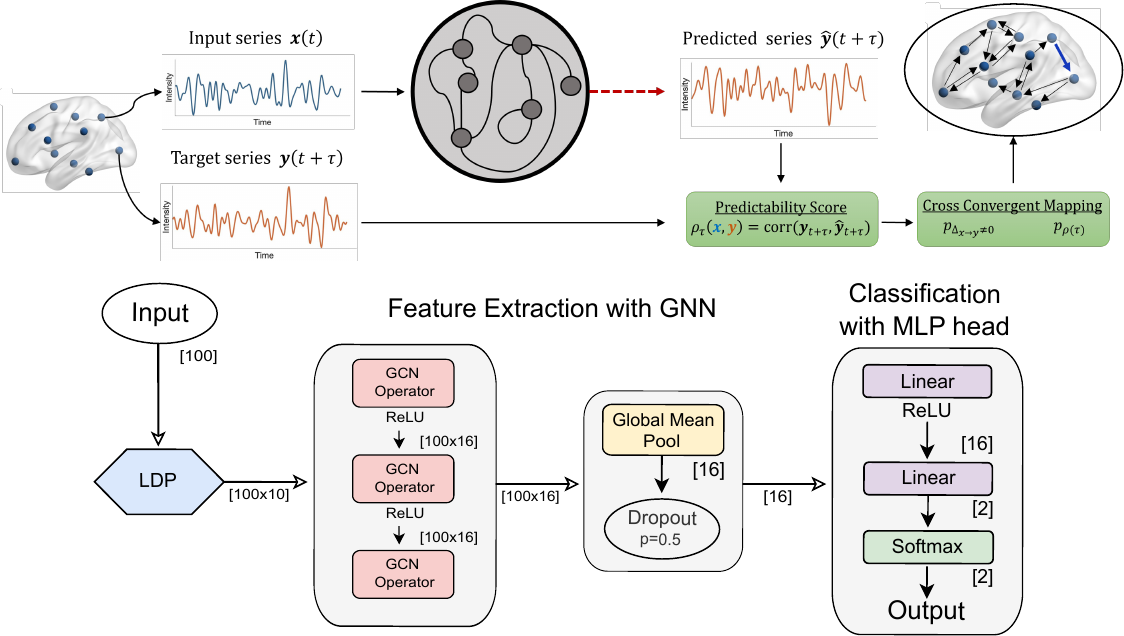}
\caption{Working diagram of causality given by the  reservoir computing (TOP) and graph convolutional architecture (BOTTOM).}
\label{fig:architecture}
\end{figure*}

The non-symmetric effective connectivity maps derived are also non-attributed, that is, there are no node features to be aggregated in Eq. \ref{layer}. Although non-attributed graphs are classifiable, they dramatically increase the problem's difficulty. Fortunately, the Local Degree Profile (LDP) method effectively decreases the challenge by setting the attributes of each node to local neighboring properties \cite{cai2018simple}. Thus, we computed the \textit{in} and \textit{out} degree of each node as well as the minimum, maximum, mean, and standard deviation of the \textit{in}/\textit{out} degree of its neighbors. This created a feature vector $\mathbf{h}_i^{(0)}$ of dimension 10 that was propagated through the directed adjacency matrix for every subject. The neural network consisted of $l=2$ hidden layers and was trained for 150 epochs with a learning rate of 0.005 to minimize the binary cross entropy between the predicted and true classes (Fig. \ref{fig:architecture} BOTTOM). The metrics were computed with a balanced class weight to account for the different number of samples in each class. The model was tested in a 10-fold cross-validation scheme and used a validation set to test for overfitting.

\subsection{Local Topology Profile}
A recent extension of the LDP attribution outlined before incorporates other local properties to the already-mentioned descriptors. This Local Topology Profile \cite{adamczyk2023strengthening} has been shown to improve the accuracy over its parent version, namely LDP. Following the original proposal, we extended the feature vector $\mathbf{h}_i^{(0)}$ with the edge betweenness centrality \cite{girvan2002community}, the overlap between node neighborhoods (i.e., Jaccard index), and the local degree score \cite{lindner2015structure}. 

However, as an attempt to further reduce the complexity of the workflow, we used the 13 LTP features with a random forest classifier of max depth 2 and a maximum number of features equal to 5. As in the GCN classifier, we used class weights to balance the dataset and used a 10-fold cross-validation scheme. The architecture used in practice is summarized in 
Figure \ref{fig:architecture}. 

\subsection{Local Interpretable Machine-Agnostic Explanations}
To explain the features allowing the classification we used the  
LIME (Local Interpretable Model-agnostic Explanations) approach. This technique explains the prediction of any classifier by learning the model locally around the prediction \cite{ribeiro2016should}. In our case, this was used to highlight the edges that contributed to the classification performance the most. LIME assigns a coefficient to each edge on the EC matrix based on the contribution to the final classification score. 

Positive values were useful in identifying the stroke group, whereas negative values were consistent in identifying the control group. The total explainability values of each ROI were calculated for both groups separately. These values were thresholded with the arbitrary threshold of 0.02 for the stroke group and -0.02 for the control group (because these directions helped the correct decisions). Edges associated with wrong decisions were not studied due to their lack of meaning in neurological terms.

\section{Results}

\subsection{Effective connectivity maps derived from Reservoir Computing}
\begin{figure}[h]
    \centering
    \includegraphics[width=\linewidth]{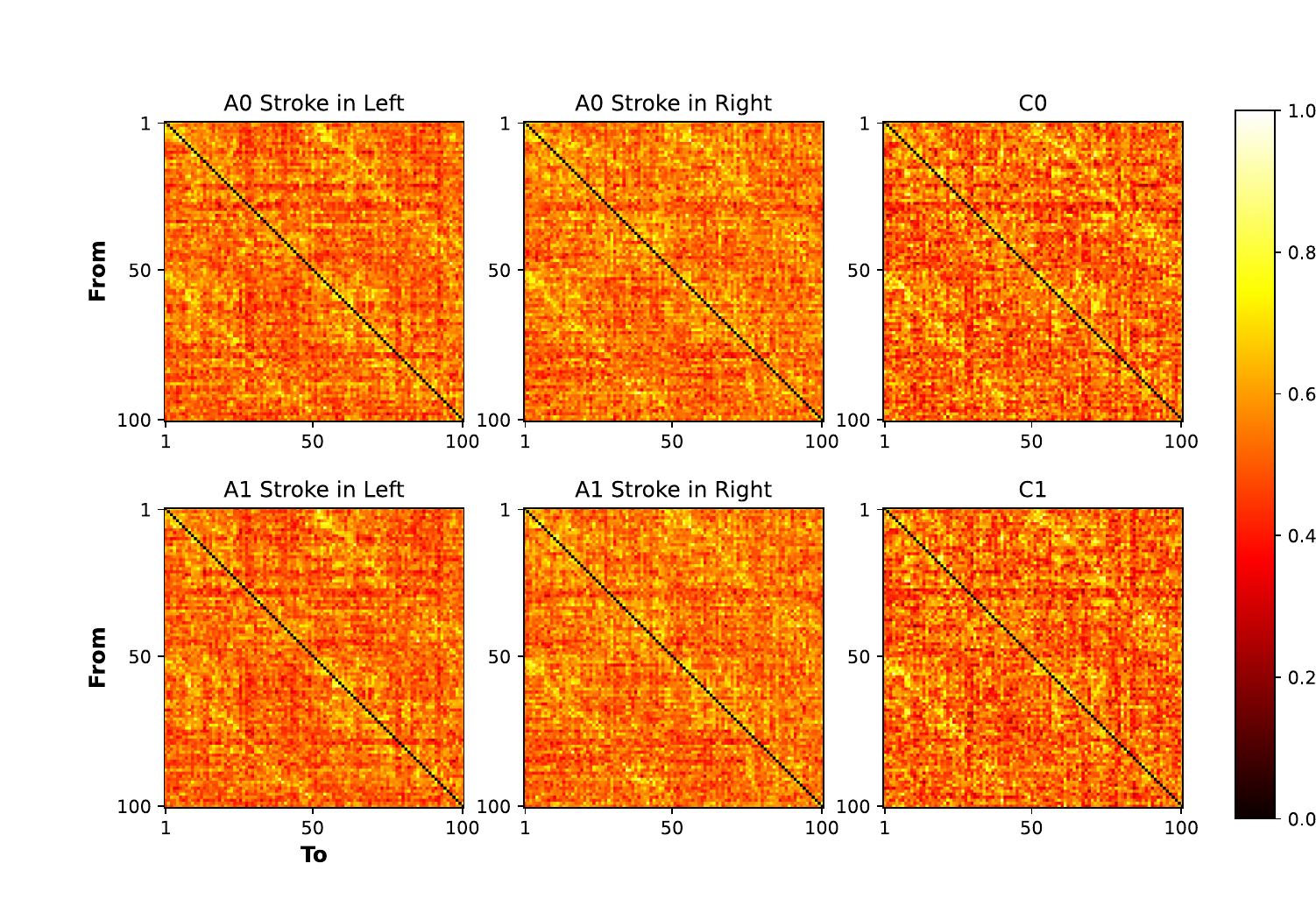}
    \caption{Group averaged effective connectivity matrices for two different Times of Repetition. Top: -1 TR. Bottom: -2 TRs. The left column is the average of subjects suffering from a stroke located in the left hemisphere. The middle column is the average of subjects suffering from a stroke located in the right hemisphere. The right column is the average of the control group.       }
    \label{fig:EC}
\end{figure}

EC maps were not readily interpretable given the complex interactions expected to occur at different spatial and temporal scales. Consensus stipulates that information transfer is obscured by the hemodynamic response function, which effectively masks the corresponding temporal delay between cause-consequence associations. We computed effective connectivity maps between 100 ROIs at two different delays (Time of Repetition = 1 and 2; see Fig. \ref{fig:EC}). The average maps showed clear patterns of hemispheric segregation while at the same time exhibiting strong connectivity between homotopic regions. In canonical functional connectivity studies, this \textit{a priori} segregated structure can be considered as an initial quality assessment of the resulting maps, forming the basis for an accurate description of the functional relationships expected to occur in brain disease.

\begin{figure}[h]
    \centering
    \includegraphics[width=\linewidth]{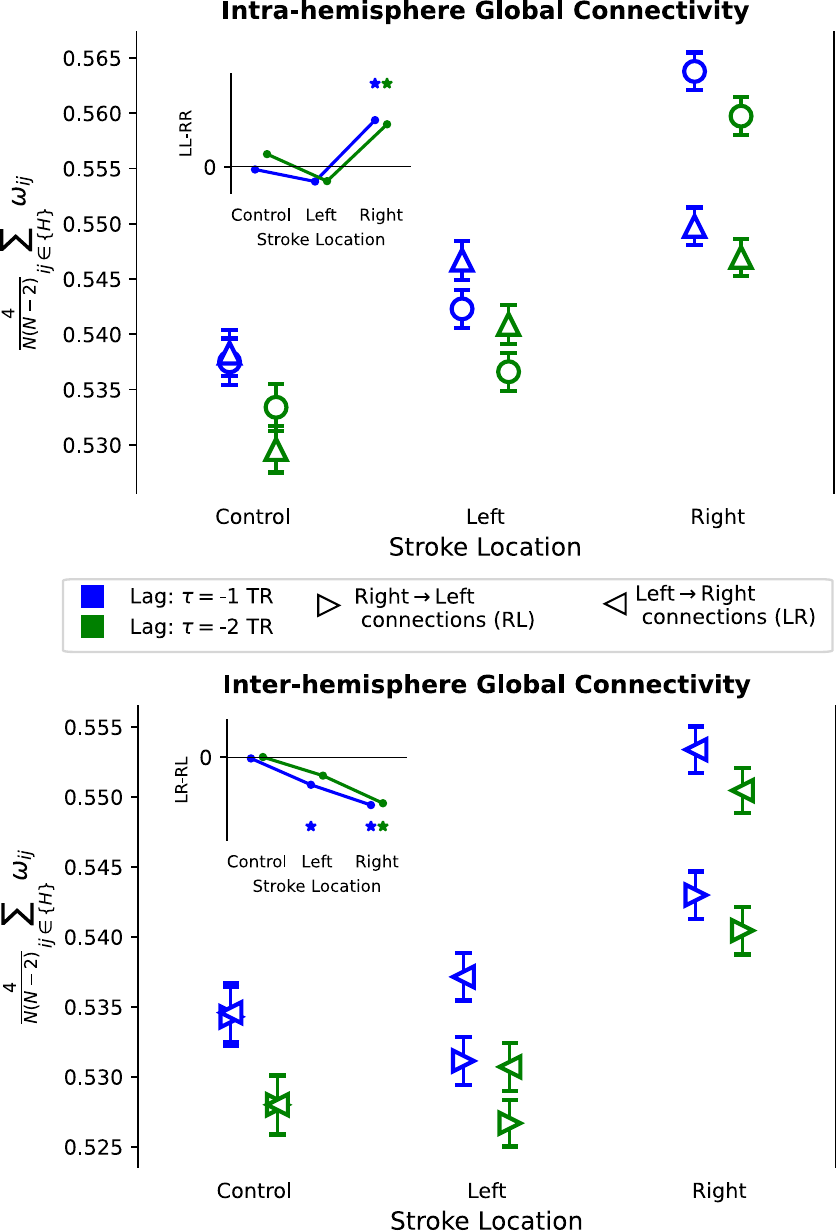}
    \caption{Global effective connectivity alterations between regions located in the same hemisphere (top) and between regions located in different hemispheres (bottom). Error bars depict 1 standard error of the mean. Insets show the average difference between left-left and right-right effective connectivity (top) and between left-right and right-left effective connectivity (bottom). Statistical significance was assessed via a two-sample t-test ('*' p<0.05). Global connectivities were obtained by averaging the weight value over the connections belonging to the corresponding hemispheres $H$.}
    \label{fig:hemispheric-EC}
\end{figure}

Even though stroke occurrence is not entirely random \cite{corbetta2015common,thiebaut2020brain}, their exact morphologies and functional disconnection patterns are highly variable. We further examined the properties of the directed networks by computing the average directed connectivity for controls, subjects suffering from right-hemispheric stroke, and subjects suffering from a stroke located on the left hemisphere (Fig. \ref{fig:hemispheric-EC}). 

Global hemispheric connectivity was computed by averaging the EC maps within and between hemispheres. That is, averaging the values in each on of the 4 visible squares in the average EC maps (Fig. 4). Briefly, intra- and inter-hemispheric connectivity was severely altered in all patients, showing a clear break of symmetric communication w.r.t. the control group, especially for right-impaired subjects \cite{Koba2024}. 

\subsection{Classification results}
The results of the classification are reported in Tables \ref{tab:stroke_performance_ltp} and \ref{tab:stroke_performance_gcn} respectively for the GCN and LTP classifiers.  
Results are reported for both the proposed method and Granger Causality: Average AUC, 
accuracy, precision, recall, and F1 are reported. As expected, the LTP (augmented with a random forest classifier) generally increased the classification metrics, although both models are comparable. It should be noted that classifying effective connectivity graphs is a complicated task due to sample heterogeneity \cite{crimi2021structurally,stroke-prognostic}, and that very similar scores compared to the chance levels (e.g., an increase of 0.2-0.3) are found in the literature \cite{adamczyk2023strengthening}. 

\begin{table}[ht]
\centering
\caption{Classification performance of the \acs{GCN} model. Results are shown by comparing the classification of \acs{EC} networks derived with the \acs{whole-brain RCC} method and the \acs{GC} method.}
\label{tab:stroke_performance_gcn}
\begin{tabular}{l | cc}
\textbf{Metric} & \textbf{whole-brain RCC} & \textbf{Granger Causality} \\
\hline
AUC score              & $0.6866 \pm 0.0830$ & $0.6074 \pm 0.0588$  \\
Accuracy                & $0.6816 \pm 0.0551$ & $0.5386 \pm 0.1610$  \\
Precision               & $0.9253 \pm 0.0654$ & $0.9178 \pm 0.0585$  \\
Recall                  & $0.6870 \pm 0.0991$ & $0.4968 \pm 0.2184$  \\
F1 score                & $0.7808 \pm 0.0511$ & $0.6143 \pm 0.1922$  \\
\end{tabular}
\end{table}

\begin{table}[ht]
\centering
\caption[Classification performance of the \acs{LTP} model.]{Classification performance of the \acs{LTP} model. Comparing the classification of \acs{EC} networks derived with the \acs{whole-brain RCC} method and the \acs{GC} method.}
\label{tab:stroke_performance_ltp}
\begin{tabular}{l | cc}
\textbf{Metric} & \textbf{whole-brain RCC} & \textbf{Granger Causality} \\
\hline
AUC score               & $0.6900 \pm 0.0652$ & $0.7240 \pm 0.1186$  \\
Accuracy                & $0.6972 \pm 0.0552$ & $0.7921 \pm 0.1377$  \\
Precision               & $0.9228 \pm 0.0523$ & $0.9121 \pm 0.0451$  \\
Recall                  & $0.7041 \pm 0.0757$ & $0.8233 \pm 0.1493$  \\
F1 score                & $0.7947 \pm 0.0443$ & $0.8606 \pm 0.1040$  \\
\end{tabular}
\end{table}

\subsection{Node and edge importance in stroke detection}
\begin{figure}[h]
    \centering
    \includegraphics[width=\linewidth]{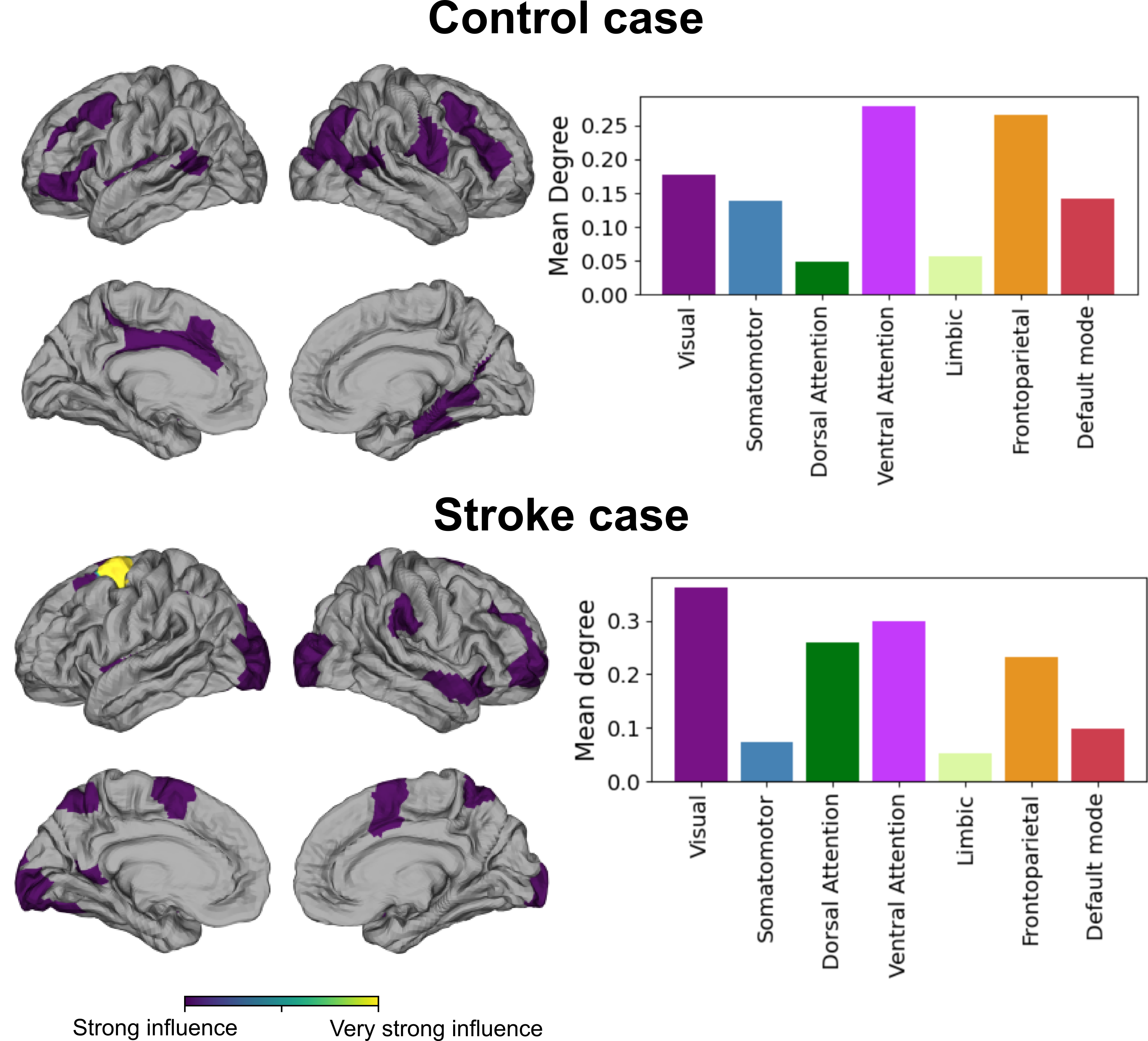}
    \caption{Interpretation of the LIME explainability outputs for each group. Cortical projection of the total contribution of each ROI (left) and its association with one of the 7 resting-state networks. The Dorsal attention network is distinctively necessary to discriminate the presence of a lesion.}
    \label{fig:ROIs}
\end{figure}

We used the LIME explainability framework on the LTP classifier due to its slightly better performance and higher computational efficiency to highlight the most descriptive ROIs and edges related to stroke onset. Importantly, the explanations were done on top of the EC matrices obtained with the reservoir method and not the granger one. For each node in the EC networks, we summed all the explainability coefficients to assess the contribution of each connection arising in each node to the correct classification (i.e., sum over all columns). Lastly, binarized and thresholded explainability values were projected back to the surface mesh (Fig. \ref{fig:ROIs}; see also Methods and \cite{Koba2024}). The resulting maps show that regions in visual, dorsal, and ventral attention have the most contribution to the classification performance for stroke subjects, while ventral attention and frontoparietal networks contributed the most to the detection of control subjects. 

\section{Discussion}
This study addresses the critical need for precise diagnostic tools in stroke management, highlighting the complexity and variability of MRI data and the limitations of conventional machine learning approaches in capturing dynamic network disruptions. The proposed pipeline begins by employing reservoir computing to define effective connectivity of the brain \cite{friston2003dynamic}. Effective connectivity using reservoir computing has been recently proposed to unravel more precise interactions in large neural systems \cite{wholebrainRCC}. However, studies that thoroughly assess the quality of the resulting causal mappings remain unseen. We propose to evaluate them by first studying existing asymetries in brain information transfer. These maps lead to directed graph representations, which have been loosely explored by graph convolutional network classifiers. Later, we used these directed maps in a AI classification and explainability paradigm; that is, disentangling regions and connections that are important for each control or stroke group. 

Functional and effective connectivity asymmetries have been previously characterized in two different formats. Using a Granger-based methodology, Allegra and colleagues \cite{allegra2021stroke} described a connectivity imbalance between lesioned and healthy hemispheres. With the maps obtained with the whole brain reservoir computing causality methodology, we observed a similar pattern which was exacerbated in subjects suffering from right-sided lesions (Figs. \ref{fig:EC} and \ref{fig:hemispheric-EC}). Furthermore, upon examining the connectivity between hemispheres, the same type of broken balance was significantly visible as well. Future work could assess how this asymmetry relates to subject behavior. With respect to this, Koba and colleagues \cite{Koba2024} explored hemispheric asymmetry in functional connectivity gradients \cite{margulies2016situating} finding a slightly higher correlation between behavior and functional aberrancy in subjects with right-sided lesions. Hence, our findings agree with the fact that the location of the stroke conveys different functional and effective information at a connectomic scale strengthening the need for a more accurate characterization of the expected behavioral dysfunctions and prognosis \cite{fox2018mapping}.

Regarding the classification paradigm, graph-structured data is ubiquitous across various disciplines, yet the use of specific graph convolutional neural networks is relatively recent (see \cite{zhou2020graph} for an extensive review). Extensions of methods for directed graph analysis have also been proposed \cite{zhang2021magnet}, modifying the architecture to perform node classification or link prediction. In this study instead, we focused on overall directed graph classification which was achieved by using conventional graph convolutions with directed adjacency matrices. We are then aggregating these Local Degrees and Topological Profiles based on the message passing across these directed connections. 

The pipeline achieves promising results, yielding an area under the curve of 0.69, superior to the state-of-art method (GC) using the GCN classification model. This should be considered a promising result given the highly heterogeneous dataset (stroke lesions were present in different parts of the brain), where similar scores relative to chance levels are often observed \cite{stroke-prognostic}. Furthermore, it was also possible to employ explainable AI tools to interpret disrupted networks despite these diversified lesions across brain networks. This elucidates the contribution of effective connectivity biomarkers that can capture aspects at a general level despite those individual differences, offering insights into disease mechanisms and treatment responses.

Previous studies on structural connectome of stroke patients highlighted network dysfunctions \cite{siegel2016effects}. Stroke-related modulations in inter- and intra-hemispheric coupling were recently investigated highlighting asymmetry
and inter-areal interactions after stroke,  related to broad changes in inter-areal communication and resulting in several deficits \cite{allegra2021stroke}.  Moreover, Erdogan and colleagues argued that the global fMRI signal is affected by the stroke lesion generating a delay of the blood-oxygen-level-dependent (BOLD) signal depending on the lesions \cite{erdougan2016correcting}. 
Our results were in line with those previous analyses. 
We found inter-hemispheric connectivity was severely altered in all patients, showing a clear break of symmetric communication w.r.t. the control group. The differences were particularly pronounced in the case of stroke lesions in the right hemisphere. This can hypothesized as the integrity of the within-hemispheric networks is sustained through language-related connections, as the right hemisphere is less involved in speech generation and suffers more from the injury. \cite{friederici2011brain}.  Indeed, the explainability maps of the control subjects resemble the vision and language networks. It is possible that the algorithm abused the connections from/to the language network to detect control subjects. Aphasia is a common symptom in the case of ischemic stroke, therefore the connections of the language network in the stroke group may show different characteristics. A similar hypothesis can be suitable also for stroke subjects because the supplementary motor area, which plays an important role in language processing, was also useful for accurate classification.  Importantly, alterations in the ventral and dorsal attention networks are often present in stroke \cite{corbetta2015common,corbetta2018low,stroke-prognostic,Koba2024}, which are in line with our explainable maps in Fig. \ref{fig:ROIs}. Nevertheless, these claims should be confirmed with larger datasets.

Undoubtedly, there are several ways to discriminate control subjects from stroke patients which are less computationally demanding   \cite{smith2017imaging}, and previous studies also showed a correlation between functional and effective connectivity with the first being easier to compute than the latter \cite{allegra2021stroke}. 
Here, we emphasized the use of a classification task for two reasons 1) to further assess the effective connectivity maps and 2) to provide a strong basis for which to implement explainability pipelines. With this we also propose an approach to classify directed graphs.  However, we showed the need to 
use further mapping into anatomical atlases to allow acceptable explainability.
Although, in conclusion, this proposes an end-to-end pipeline for studying
effective connectivity brain disorders, capitalizing on a specific approach for directed graph and explainability.  

This   analytical framework  enhances clinical interpretability but also can inspire confidence in decision-making processes, crucial for translating research findings into clinical practice as it can translate complex neuroimaging features into simple visualizations. The study lays the groundwork for improved patient stratification in other brain diseases as well, with the ultimate goal of  assisting doctors,  demonstrating also the potential of reservoir computing causality, graph convolutional networks, and explainable analysis.

\section*{Acknowledgement}
Authors thank Prof. Maurizio Corbetta
for sharing the dataset used in this study.  The publication was created within the project of the Minister of Science and Higher Education "Support for the activity of Centers of Excellence established in Poland under Horizon 2020" on the basis of the contract number MEiN/2023/DIR/3796. 
This project has received funding from the European Union’s Horizon 2020 research and innovation programme under grant agreement No 857533. 
This publication is supported by Sano project carried out within the International Research Agendas programme of the Foundation for Polish Science, co-financed by the European Union under the European Regional Development Fund. This research was supported in part by the PLGrid infrastructure. 
Computations have been partially performed on the ARES supercomputer at ACC Cyfronet AGH.

\nolinenumbers

\bibliography{main}

\bibliographystyle{abbrv}

\end{document}